\documentclass{amii}

\usepackage{listings}
\usepackage{mathtools}
\usepackage{siunitx}
\usepackage{xspace}
\usepackage{ntheorem}
\usepackage{bbm}
\usepackage{stmaryrd}
\usepackage{nth}
\usepackage{tikz}
\usepackage{pgfplots}
\usepackage{multirow}
\usepackage{placeins}
\usepackage{soul}
\usepackage[linesnumbered,ruled,vlined]{algorithm2e}
\usepackage{algpseudocode}
\SetKwInput{KwInput}{Input}
\usepackage{lipsum}
\usepackage{seqsplit}
\usepackage{multicol}
\usepackage{adjustbox}
\usepackage[normalem]{ulem}
\useunder{\uline}{\ul}{}
\usepackage{paralist}
\newcommand{\fullset}{\mathcal{D}}
\newcommand{\taskset}{\mathcal{D}_{\text{dev}}}
\newcommand{\selset}{\mathcal{D}_{\text{sel}}}
\newcommand{\testset}{\mathcal{D}_{\text{test}}}

\newcommand{\Nsel}{N_{\text{sel}}}

\newcommand{\params}{\boldsymbol{\theta}}

\newcommand{\seldist}{P_{\text{sel}}}
\newcommand{\taskdist}{P_{\text{dev}}}
\newcommand{\method}{{ICLHerd}}

\title{ ICLHerd: Herding-Based Data Selection for LLMs with In-Context Learning as a Proxy}

\author[1]{Yuqiao Wen}
\author[1,2]{Lili Mou}

\affiliation[1]{Dept. Computing Science \& Alberta Machine Intelligence Institute (Amii), University of Alberta}
\affiliation[2]{Canada CIFAR AI Chair}

\metadata[Emails]{\email{yq.when@gmail.com}\quad\email{doublepower.mou@gmail.com}}

\abstract{Data selection plays a central role in modern large language model training, as the quality of training data strongly influences downstream performance.
A common approach is to curate a development set that encodes the desired skills (e.g., reasoning or multilinguality) and select data that are similar to them.
Existing methods compute similarity using cosine distance over either the model's hidden states or gradients.
However, we observe that this max-similarity strategy induces a mode-seeking bias: the selected data clusters tightly around the development samples.
To address this, we propose \method{}, a data selection framework that leverages the von Mises--Fisher kernel to balance mode-seeking with mode-averaging.
We further introduce in-context learning as a validation proxy, enabling efficient hyperparameter search over the balance trade-off.
Results show that \method{} outperforms existing LLM data selection algorithms, highlighting the importance of balancing mode-seeking and mode-averaging.}

\usepackage{amsmath,amsfonts,bm,xspace,multirow}
\definecolor{myred}{RGB}{255,127,127}
\definecolor{mygreen}{RGB}{127,255,127}

\def\1{\bm{1}}

\DeclareMathAlphabet{\mathsfit}{\encodingdefault}{\sfdefault}{m}{sl}
\SetMathAlphabet{\mathsfit}{bold}{\encodingdefault}{\sfdefault}{bx}{n}

\newcommand{\E}{\mathop{\mathbb{E}}}

\theoremstyle{plain}

\begin{document}

\maketitle

\section{Introduction}
\label{sec:intro}

Data selection is the task of choosing training samples from a data pool in order to maximize model performance given a budget of training data.
This is especially important in the large language model (LLM) era, where the candidate data pool is potentially the entire Internet~\citep{wenzek2020ccnet, penedo2024fineweb, li2024datacomp}.
In modern LLM training, data selection is usually one of the most crucial steps, 
and researchers rely heavily on intuition and trial-and-error to curate an appropriate training set tailored for the desired model capability.
For example, DeepSeek-V4 incorporated different filtering strategies to remove templated data (e.g., repetitive html scripts) and mitigate the risk of model collapse~\citep{xu2026deepseek};
Qwen 3 was trained across a selected set of 119 languages and dialects in an attempt to improve data diversity and improve the model's multilingual skills~\citep{yang2025qwen3};
Llama 4 was trained on what the technical report claimed as ``specialized datasets'' to improve the model's long-context capabilities~\citep{llama4}.
It is important to note that the training recipes of these frontier open-weight models are not publicly available.
This leaves a gap in systematic research on data selection,
as the scale of training data (often trillions of tokens) makes it infeasible to ablate individual selection strategies,
limiting the scientific rigor of these technical reports.

Existing open research on LLM data selection generally uses a development (dev) set to guide the selection process, where the dev samples encode the desired capabilities, e.g., math and reasoning.
A promising direction is to embed all the samples and select training data based on the similarity to the dev set.
\citet{xia2024less} propose LESS, which embeds training samples via down-projected LoRA gradients, and at each iteration, selects the training sample that maximizes cosine similarity to any dev sample.
\citet{ivison2025large} propose RDS+, which embeds samples using the model's own hidden states for efficiency,
and employs a round-robin selection that cycles through dev samples in fixed order, matching each dev sample in turn to promote data diversity.

Despite the efforts, we observe that these algorithms suffer from the \textit{mode-seeking} problem, where the selected dataset is overly crowded around the dev samples.
Here, a \textit{mode} refers to a peak in the embedding space, and both algorithms tend to select samples that are close to one or a few peaks of the dev distribution: 
LESS greedily pick the sample that match well against any of the dev samples, whereas RDS+ attempt to match all dev samples in a round-robin fashion.
Alternatively, we may also select data based on the mean embedding of the dev set, as opposed to the maximum used in LESS and RDS+.
This, on the other hand, will lead to the \textit{mode-averaging} problem, where the embeddings of the selected dataset will center around the centroid of the dev distribution, losing the diversity of capabilities encoded by different dev samples.

To this end, we propose \method, a data selection framework based on the principled data herding scheme~\citep{welling2009herding, chen2010super}, where LESS and RDS+ emerge as special cases. The general idea is to make the distribution of selected samples close to that of the dev set. We use the maximum mean discrepancy~\citep[MMD,][]{gretton2012kernel} to measure the similarity between two distribution based on the selected samples and those in the dev set. In particular, we adopt the von Mises--Fisher (vMF) kernel~\citep{vmf1987} on normalized features as the similarity measure, which operates on cosine similarities, since LLM representations are commonly compared by their direction rather than their magnitude~\citep{banerjee2005clustering, reimers2019sentence, gao2021simcse}.
We also design a fused kernel that combines hidden-state and gradient representations, allowing us to leverage their complementary signals for data selection.
Overall, the data selection reduces to the problem of looking for similar samples with a diversity penalty. We apply the herding algorithm~\citep{welling2009herding, chen2010super} for optimization, which greedily selects the best sample at a time and adjusts the diversity penalty accordingly.

To efficiently tune the balance between mode averaging and mode seeking, we further propose in-context-learning-as-a-proxy for our \method{} framework.
Typically, hyperparameter tuning on data selection requires repeatedly training the model and validating the model's performance, which can be prohibitively expensive.
Our proposed \method{} circumvents this problem by validating through in-context learning, where we put the selected data as a prompt and evaluate its loss on the dev set. Note that this is especially efficient due to prefix caching, as the same prompt is shared among dev samples. Our results establish a connection between in-context learning and gradient-descent-based learning performance.

Overall, we propose a \method{} framework of data selection for training LLMs. Our contributions are three-fold:
(1) we address the mode-averaging and mode-seeking problems in data selection for LLMs;
(2) we design a fused vMF kernel combining hidden states and gradients to control this balance; and
(3) we propose in-context-learning-as-a-proxy for efficiently tuning it.

\section{Related Work}
\label{sec:related}

Data selection for language model training aims to choose a subset of a large candidate pool to reduce training costs~\citep{albalak2024survey, qin2024unleashing}. Existing methods span different paradigms.

\textbf{Classifier-based data filtering.}
Classifier-based methods train a lightweight classifier to separate desired samples from the raw pool, and select samples by a threshold on the classifier score, with early studies in machine learning focusing on identifying mislabeled examples and eliminating noise~\citep{wilson1972asymptotic, brodley1999identifying}.
For LLM training, the classifiers are usually trained with high-quality text as positive data, e.g., Wikipedia~\citep{brown2020language, gao2020pile}.
Alternatively, positive samples can also be task-specific data, such as instruction following~\citep{xie2023data, li2024datacomp}.
More recently, researchers also use an LLM to annotate positive samples and ensemble several classifiers for different aspects of data quality~\citep{penedo2024fineweb, su2025nemotron}.
While these methods are efficient enough for pretraining pools, they score each document independently and do not consider the distribution of the selected set.

\textbf{Loss and perplexity scoring.}
Another family of data selection methods filter examples by a language model's loss or perplexity, treating this as a proxy for sample utility~\citep{lewis1994sequential, moore2010intelligent, wenzek2020ccnet}.
Existing approaches for LLM training differ on which part of the loss range to preserve: the hardest examples, the middle~\citep{marion2023less, ankner2024perplexed}, or the easiest, such as the most likely response to each instruction~\citep{zhang2025best}.
For instruction tuning, \citet{li2024from} compare the loss of the response with and without the instruction, ensuring that difficulty reflects the instruction rather than the response alone.
However, a loss score only measures how difficult an example is for the model; since a useful example can be either easy or hard, such methods may overlook other aspects of data quality~\citep{wettig2024qurating, sachdeva2026train}.

\textbf{Similarity-based data filtering.}
Similarity-based methods select data that are most similar to a small dev set from the target task~\citep{shimodaira2000improving, huang2006correcting, koh2017understanding}.
Specifically, they embed the pool and dev samples into a representation space and compute similarity between them.
The choice of representation varies, including the gradient of the model's loss~\citep{xia2024less, dai2025improving}, the hidden states of the model being trained~\citep{ivison2025large}, or the output of a separate embedding model~\citep{ivison2025large, yin2025compute}.
They also differ in their selection scheme:
\citet{xia2024less} rank examples by their best match to any dev sample;
\citet{ivison2025large} go through the dev samples in a round-robin fashion to equally optimize for every dev sample;
and \citet{dai2025improving} iteratively select the sample that most benefits the currently most underrepresented task.
Despite these efforts, there is a lack of systematic study on which features and which selection scheme to use.
These similarity-based methods also rely on the best matches to the dev samples, whereas our \method{} selects a set of samples whose distribution is close to that of the dev set.

\textbf{LLM-as-a-judge for data filtering.}
LLM-as-a-judge (LaaJ) uses a strong language model to evaluate text~\citep{zheng2023judging, liu2023geval, gu2024survey},
and has also been used for data selection by rating each training example and keeping the high-scoring ones~\citep{chen2024alpagasus, wettig2024qurating, sachdeva2026train}.
These methods differ in how they define quality, e.g., accuracy~\citep{chen2024alpagasus}, complexity~\citep{liu2023makes}, and multi-criteria rubrics~\citep{wettig2024qurating}.
Researchers also develop different ways to elicit a judgment, e.g., a 0--5 rating~\citep{chen2024alpagasus}, the probability of a ``yes'' answer~\citep{sachdeva2026train}, and pairwise comparison~\citep{wettig2024qurating}.
However, scoring a whole pool with a large judge is computationally expensive; to address this, some methods distill judge labels into a small trained scorer~\citep{liu2023makes, wettig2024qurating}.
Since each example is scored independently and may be redundant, a separate mechanism for coverage is often added to the approach, such as a diversity filter and a sampling temperature~\citep{liu2023makes, wettig2024qurating, sachdeva2026train}.

Recent work shows that random selection and hidden-state similarity are {strong baselines} for LLMs~\citep{diddee2024chasing, ivison2025large}, a finding consistent with our results.
This suggests that data selection for LLMs is still challenging and warrants further investigation.

\section{Method}
\label{sec:method}

\subsection{Problem Formulation}

Consider a standard train, dev, and test split of data, where we have three distinct sets:
$\fullset$ for model training, $\taskset$ for hyperparameter tuning, and $\testset$ for model evaluation.
In the context of LLMs, a model is usually pretrained with massive unlabeled data and then finetuned with in-domain data to improve performance. Even for fine-tuning, the training set $\fullset$ is often a gigantic pool of unfiltered texts scraped from the Internet, most of which are noisy and out of distribution with respect to $\taskset$ and $\testset$~\citep{albalak2024survey, penedo2024fineweb, qin2024unleashing}.

In this paper, we study data selection: our goal is to filter $\fullset$ to a much smaller subset, where the selected data are better aligned to the task that we care about, for example, reasoning and instruction following.
Under this formulation, we utilize $\taskset$ to guide the selection process, in addition to its role of hyperparameter tuning in standard machine learning. 
Formally, given a budget of $\Nsel$ training samples, our goal is to find the optimal subset $\selset$ that minimizes a loss on the dev set
\begin{align}
    &\operatorname*{minimize}_{\selset \subset \fullset} \; \mathcal{L}\!\left( \taskset;\; \operatorname*{argmin}\limits_{\params} \mathcal{L}(\selset;\params) \right) \quad \text{subject to} \quad |\selset| = \Nsel,
\end{align}
where $\mathcal L(\cdot; \params)$ computes the loss on a dataset with parameter $\params$, and $\operatorname*{argmin}_{\params} \mathcal{L}(\selset;\params)$ is typically given by a gradient descent algorithm as an approximation.

Note that this optimization problem is difficult,
because it is an integer programming problem with non-linear loss function that involves an expensive $\operatorname*{argmin}$ operator over the entire parameter space.
Brute forcing this problem requires training the model $\binom{|\fullset|}{\Nsel}$ times and choosing the best subset, which is not feasible in practice.

\subsection{Our \method{} Approach}

The overall goal of our approach is to select a dataset $\selset$, whose distribution is close to that of $\taskset$. However, we only have discrete data samples in $\selset$ and $\taskset$, without an explicit form of their distributions. Thus, we use the maximum mean discrepancy~\citep{gretton2012kernel} to measure distribution similarity based on samples using a kernel function. With such a formulation, the objective becomes selecting samples that are close to $\taskset$, but with a diversity penalty. We minimize this objective by the classic herding algorithm, which selects samples greedily while maintaining a running profile of diversity scores.

In the rest of this section, we will first introduce the MMD formulation, kernel design, and the herding algorithm. Then, we will present an in-context-learning-as-a-proxy method for efficient hyperparameter tuning in the LLM setting.

\textbf{MMD formulation.} We assume $\selset$ and $\taskset$ are sampled from two underlying distributions $\seldist$ and $\taskdist$,
respectively.
Specifically, the MMD between $\seldist$ and $\taskdist$ is defined as
\begin{align}
    \operatorname{MMD}(\seldist,\taskdist) = \big\| \E_{\mathbf x \sim \seldist}[\phi(\mathbf x)] - \E_{\mathbf y \sim \taskdist}[\phi(\mathbf y)] \big\|_{\mathcal H}^2,
\end{align}
where $\phi:\mathcal X \to \mathcal H$ is a feature mapping into a reproducing kernel Hilbert space $\mathcal H$,
and  $\|\cdot\|_{\mathcal H}$ is the norm in the Hilbert space. We estimate the MMD using finite datasets $\selset$ and $\taskset$, given by
{\medmuskip=0mu \thickmuskip=1mu \thinmuskip=1mu \nulldelimiterspace=0pt
\begin{align}
    {\operatorname{MMD}} \approx \Big\| \frac{1}{|\selset|} \sum_{\mathbf x \in \selset} \phi(\mathbf x) - \frac{1}{|\taskset|} \sum_{\mathbf y \in \taskset} \phi(\mathbf y)  \Big\|^2.
\end{align}}When expanding the formula, we replace $\phi(\mathbf \cdot)^\top\phi(\cdot)$ with a general kernel function $k(\cdot, \cdot)$. This allows us to consider sample similarity without explicitly defining $\phi(\cdot)$, as $k(\cdot, \cdot)$ serves as an inner product in the $\phi(\cdot)$ space.

We further drop the $\frac{1}{|\taskset|^2}\sum_{\mathbf y, \mathbf y' \in \taskset} k(\mathbf y, \mathbf y')$ term, as it is independent of $\selset$. The objective becomes
\begin{align}
    \operatorname*{minimize}_{\selset \subset \fullset} {\frac{1}{|\selset|^2}} \sum_{\mathbf x, \mathbf x' \in \selset} k(\mathbf x, \mathbf x') - {\frac{2}{|\selset||\taskset|}} \sum_{\mathbf x \in \selset, \mathbf y \in \taskset} k(\mathbf x, \mathbf y).
\end{align}

We also convert this to a maximization problem for clarity, and introduce a hyperparameter $\alpha$ balancing the two terms instead of using fixed coefficients. Overall, we have the following objective
\begin{align}
    \operatorname*{maximize}_{\selset \subset \fullset} \mkern 12mu \underbrace{\sum_{\mathclap{\mathbf x \in \selset, \mathbf y \in \taskset}} k(\mathbf x, \mathbf y)}_{\text{similarity}} \;-\; \alpha \underbrace{\sum_{\mathclap{\mathbf x, \mathbf x' \in \selset}} k(\mathbf x, \mathbf x')}_{\text{diversity penalty}}.
    \label{eqn:mmd}
\end{align}

Intuitively, the first term aims to maximize the similarity with $\taskset$, the second term encourages selecting diverse data samples, avoiding the selected samples crowding around certain modes. Note that the diversity objective is naturally derived from the MMD formulation, whereas previous work uses heuristics to select diverse data~\citep{ivison2025large}.

\textbf{Kernel design.}
A common choice of the kernel is the radial basis function (RBF).
For neural networks, the von Mises--Fisher kernel~\citep{vmf1987} is often preferred~\citep{kobayashi-2018-frustratingly}, as the representations are often compared by their direction rather than the Euclidean distance.
We empirically analyze different kernels and observe that vMF is indeed superior to other options.

Specifically, the von Mises--Fisher (vMF) kernel scales exponentially with the cosine similarity controlled by a concentration parameter.
This allows us to tune how sharply similarity decays as samples move away from the dev samples.
Formally, we have
\begin{align}
    k_{\text{vMF}}(\mathbf x, \mathbf y) = \exp\big(\kappa\, \psi(\mathbf x)^\top \psi(\mathbf y)\big),
    \label{eqn:vmf}
\end{align}
where $\psi(\cdot)$ is a unit-norm feature map that represents the samples in an embedding space, and $\kappa$ is the concentration parameter that balances between mode-averaging (mean similarity) and mode-seeking (max similarity).

As shown in~ Eqn.~(\ref{eqn:vmf}), a high $\kappa$ causes mode seeking, since the influence of distant samples diminishes rapidly,
encouraging the selected subset to cluster around the development samples.
On the other hand, a low $\kappa$ causes mode averaging, as it places greater emphasis on preserving global similarity and results in selected samples crowded at the center of the dev samples (which may not resemble any dev samples).

For the representation $\psi(\cdot)$ of a sample, previous studies use either gradients or hidden states of an LLM~\citep{xia2024less, ivison2025large}, but we notice that they complement each other.
Therefore, we propose a fused vMF kernel that utilizes signals from both hidden states and gradients
\begin{align}
    k_{\text{fused-vMF}}(\mathbf x, \mathbf y) = \beta \, k_{\text{vMF}}\big(\psi_{\text{emb}}(\mathbf x), \psi_{\text{emb}}(\mathbf y)\big) + (1 - \beta) \, k_{\text{vMF}}\big(\psi_{\text{grad}}(\mathbf x), \psi_{\text{grad}}(\mathbf y)\big)
    \label{eqn:greedy-vmf}
\end{align}
where $\psi_\text{grad}(\cdot)$ and $\psi_\text{emb}(\cdot)$ compute the unit-norm feature map based on the model's gradients and hidden states, respectively, and $\beta$ controls the strength of each signal.
Specifically, we follow \citet{xia2024less}  to down-project the gradients, which reduces the memory cost, and follow \citet{ivison2025large}  to pool the hidden states over tokens.

\textbf{Herding algorithm.} We adopt the herding algorithm~\citep{bach2012equivalence} to optimize the MMD objective.
Essentially, data herding is a greedy algorithm that chooses samples to maximize Eqn.~(\ref{eqn:mmd}) iteratively while keeping a running selected set for the diversity penalty. 

First, the selected set is initialized as $S_0 = \emptyset$.

For each step $t$, we pick one sample $\mathbf x_t$ by
\begin{align}
\operatorname*{argmax}_{\mathbf x \in \fullset \setminus S_{t-1}} \Bigg[\sum_{\mathbf y \in \taskset} k(\mathbf x, \mathbf y) - \alpha\sum_{\mathbf x' \in S_{t-1}} k(\mathbf x, \mathbf x')\Bigg]
\end{align}
The selected sample $\mathbf x_t$ is added to the selected set, given by
$S_t = S_{t-1} \cup \{\mathbf x_t\}$.

Overall, our \method{} generalizes existing similarity-based methods and better balances between extreme mode averaging and mode seeking.

\textbf{In-context learning proxy.} \method{} has several hyperparameters, such as the concentration parameter $\kappa$ that balances mode averaging and mode seeking, the diversity weight $\alpha$, and the fusion weight $\beta$, which need to be tuned properly.
Traditional hyperparameter tuning for data selection requires repeated model training and subsequent inference to evaluate the selected subset, which is computationally expensive and limits how thoroughly we can balance between mode-seeking and mode-averaging.

To this end, we propose in-context learning as an efficient validation proxy, where we place the selected data samples in the model's context and measure the loss on the dev set to gauge the data quality.
Intuitively, higher-quality training samples are also more informative in-context learning samples, thus resulting in a lower loss on the dev set when they are used to prompt an LLM. This allows us to estimate the quality of the data without training on the dataset.

However, it is noisy and still costly to perform in-context learning evaluation for each individual samples. We propose to estimate the quality of $\selset$ by repeatedly drawing $k$ samples and evaluate their quality.

Let $\mathbf x_1, \mathbf x_2, \ldots, \mathbf x_k \in \selset$ be $k$ samples randomly drawn from the selected dataset such that it fits in a budget of $B$ tokens available for in-context learning.
We build an in-context learning prompt by concatenation:
\begin{align}
    \mathbf c = \mathbf x_1 \oplus \mathbf x_2 \oplus \cdots \oplus \mathbf x_k.
    \label{eqn:context}
\end{align}

We then estimate quality of the in-context learning samples $\mathbf c$ by computing the loss on the dev set, given by
\begin{align}
    \mathcal{L} (\mathbf c) = \frac{1}{|\taskset|} \sum_{\mathbf x \in \taskset} \bigg( -\frac{1}{|\mathbf x|} \sum_{t=1}^{|\mathbf x|} \log p_{\params}\big(x_t \mid \mathbf c, \mathbf x_{<t}\big) \bigg).
    \label{eqn:icl-per-draw}
\end{align}

Although one subset $\mathbf x_1, \mathbf x_2, \ldots, \mathbf x_k$ already gives a decent proxy for data quality, this score may be susceptible to random noise.
To mitigate this, we repeat this process $M$ times with different random samples, each forming a context $\mathbf c_m$ for $m=1,\cdots, M$.

Finally, we aggregate a proxy score as the maximum loss across different draws
\begin{align}
    \operatorname{ICLProxy}(\selset) = \max_{m=1}^M \mathcal{L}(\mathbf c_m)
    \label{eqn:icl-proxy}
\end{align}
Note that it is possible to also take the average instead the maximum, but our pilot study suggests that max aggregation is more consistent with actual data quality, potentially because bad outlier samples are especially detrimental to training.

\method{} is efficient, as we use a prefix caching technique when evaluating different samples in $\taskset$. In other words, the cached prefix $\mathbf c$ is computed once and reused for different dev samples, so each dev sample only requires processing its own tokens. Overall, our proposed method avoids repeated LLM training for hyperparameter tuning and is much faster than the traditional training--validation framework.

\section{Experiments}
\label{sec:experiments}

\subsection{Setups}
\label{sec:setup}

\textbf{Selection pool dataset.}
Following \citet{ivison2025large}, we use the unfiltered T\"ulu 2 SFT mixture dataset~\citep{ivison2023camels} as our selection pool $\fullset$, containing approximately 5.8M samples from sources including Open Orca~\citep{mukherjee2023orca}, FLAN~\citep{chung2024scaling}, and WizardLM Evol-Instruct~\citep{xu2024wizardlm}.
This pool combines data across various skills, requiring the selection algorithm to identify examples relevant to each target task.

\begin{table}[t]
\centering
\caption{Main results on five tasks with 10K selected samples per task.
Training data are selected using Qwen3-1.7B representations and used to finetune Qwen3-1.7B, 4B, and 8B.
Due to resource constraints, Top-PPL and Mid-PPL are only evaluated at 1.7B as their performance is relatively low.
The best and second-best results within each model size are \textbf{bold} and \underline{underlined}, respectively.
\method{} significantly outperforms RDS+, the strongest baseline, under a two-sided binomial test across tasks and model sizes ($p \approx 0.007$).}
\label{tab:main-results}
\small
\setlength{\tabcolsep}{4pt}
\begin{tabular}{|c|l|S[table-format=2.2]S[table-format=2.2]S[table-format=2.2]S[table-format=2.2]S[table-format=2.2]|S[table-format=2.2]|}
\hline
\multicolumn{8}{|c|}{\textit{Selected on Qwen3-1.7B, fine-tuned on Qwen3-1.7B}} \\
\hline
Size & Method & {GSM8K} & {MLQA} & {DROP} & {XSum} & {OASST1} & {Avg} \\ \hline
\multirow{7}{*}{1.7B} & Random & 40.33 & 48.27 & 8.13 & 15.31 & 20.37 & 26.48 \\
 & Top-PPL~\citep{ankner2024perplexed} & 13.57 & {\underline{51.63}} & {\textbf{16.72}} & 6.24 & 4.71 & 18.57 \\
 & Mid-PPL~\citep{marion2023less} & 42.08 & 21.30 & 9.06 & 13.60 & 19.77 & 21.16 \\
 & LESS~\citep{xia2024less} & 42.53 & 38.47 & 9.03 & 22.58 & 21.06 & 26.73 \\
 & RDS+~\citep{ivison2025large} & {\underline{58.91}} & 51.31 & 7.84 & {\underline{24.47}} & {\underline{21.41}} & {\underline{32.79}} \\
\cline{2-8}
 & \method{} (ours) & {\textbf{59.14}} & {\textbf{60.61}} & {\underline{14.69}} & {\textbf{24.66}} & {\textbf{21.67}} & {\textbf{36.15}} \\
\hline \hline
\multicolumn{8}{|c|}{\textit{Selected on Qwen3-1.7B, fine-tuned on other Qwen3 models}} \\
\hline
Size & Method & {GSM8K} & {MLQA} & {DROP} & {XSum} & {OASST1} & {Avg} \\ \hline
\multirow{4}{*}{4B} & Random & 50.80 & {\underline{30.85}} & 12.72 & 18.24 & 22.49 & 27.02 \\
 & LESS~\citep{xia2024less} & 65.73 & 29.96 & {\underline{16.40}} & 26.25 & {\textbf{23.22}} & {\underline{32.31}} \\
 & RDS+~\citep{ivison2025large} & {\underline{73.31}} & 19.00 & 14.07 & {\underline{27.13}} & {\underline{23.02}} & 31.31 \\
 & \method{} (ours) & {\textbf{76.57}} & {\textbf{39.55}} & {\textbf{25.18}} & {\textbf{27.49}} & 22.46 & {\textbf{38.25}} \\
\hline
\multirow{4}{*}{8B} & Random & 69.29 & 49.22 & 14.76 & 14.08 & 23.67 & 34.20 \\
 & LESS~\citep{xia2024less} & 66.41 & 54.79 & {\underline{15.71}} & 27.42 & {\underline{24.05}} & 37.68 \\
 & RDS+~\citep{ivison2025large} & {\textbf{79.83}} & {\underline{55.22}} & 13.60 & {\underline{28.72}} & 23.71 & {\underline{40.22}} \\
 & \method{} (ours) & {\underline{76.27}} & {\textbf{57.56}} & {\textbf{17.47}} & {\textbf{29.19}} & {\textbf{24.12}} & {\textbf{40.92}} \\
\hline
\end{tabular}
\end{table}

\textbf{Evaluation tasks and datasets.}
We evaluate all methods on five tasks in a zero-shot setting: GSM8K~\citep{cobbe2021training}, MLQA~\citep{lewis2020mlqa}, DROP~\citep{dua2019drop}, XSum~\citep{narayan2018dont}, and OASST1~\citep{kopf2023openassistant}.
These tasks cover a diverse range of skills, including mathematical reasoning, multilingual question answering, reading comprehension, summarization, and instruction following. For in-depth analyses, we focus on GSM8K and MLQA to test two distinct capabilities: reasoning and multilinguality. Details are presented in Appendix~\ref{app:implementation}.

\textbf{Baselines.}
We compare \method{} against five representative baselines spanning random, loss-based, and representation-based selection.

$\bullet$ \underline{Random}: uniform sampling from the pool.
This is a surprisingly strong baseline that often remains competitive~\citep{diddee2024chasing, ivison2025large}.

$\bullet$ \underline{Top-PPL}~\citep{ankner2024perplexed}: selecting samples with the highest perplexity loss under the base model.

$\bullet$ \underline{Mid-PPL}~\citep{marion2023less}: selecting samples around the median perplexity loss under the base model.

$\bullet$ \underline{LESS}~\citep{xia2024less}: embedding samples via gradients and ranking them by maximum cosine similarity to any dev sample.

$\bullet$ \underline{RDS+}~\citep{ivison2025large}: embedding samples via hidden states and selecting the best match for each dev sample in a round-robin fashion.

We provide additional details on the datasets in Appendix~\ref{app:dataset-stats}.

\subsection{Results and Analyses}
\label{sec:results}

\begin{table}[t]
\centering
\caption{Analysis of the concentration parameter $\kappa$ using the fusion kernel ($\beta=0.5$, no diversity term).
Mean- and max-similarity on concatenated features serve as the approximate extremes for small and large $\kappa$, respectively.
Random and round-robin (on fused features) lie outside the $\kappa$ continuum and are shown as references.
$\kappa$ is given as a multiple of the value from the median heuristic~\citep{gretton2012kernel}.}
\label{tab:kappa-sweep}
\small
\setlength{\tabcolsep}{3pt}
\begin{tabular}{|l|S[table-format=2.2]|S[table-format=2.2]|S[table-format=2.2]S[table-format=2.2]S[table-format=2.2]S[table-format=2.2]S[table-format=2.2]S[table-format=2.2]|S[table-format=2.2]|S[table-format=2.2]|}
\hline
{\multirow{2}{*}{Task}} & {\multirow{2}{*}{Random}} & {\multirow{2}{*}{Mean-similarity}} & \multicolumn{6}{c|}{\method{} (ours), $\kappa$} & {\multirow{2}{*}{Max-similarity}} & {\multirow{2}{*}{Round-robin}} \\ \cline{4-9}
 & & & {0.01$\times$} & {1$\times$} & {3$\times$} & {10$\times$} & {30$\times$} & {100$\times$} & & \\ \hline \hline
GSM8K & 40.33 & 54.28 & 56.56 & 56.25 & {\underline{57.47}} & {\textbf{58.83}} & 55.88 & 55.88 & 54.74 & 53.53 \\
MLQA & 48.27 & {\underline{60.34}} & {\textbf{60.84}} & 59.99 & 59.71 & 43.13 & 44.63 & 45.29 & 47.20 & 59.31 \\
\hline
\end{tabular}
\end{table}

\begin{table}[t]
\centering
\caption{Comparison of feature spaces.
RDS+ and LESS use hidden states and gradients, respectively, while \method{} (fused) combines both in the kernel.}
\label{tab:feature-space}
\small
\begin{tabular}{|l|S[table-format=2.2]S[table-format=2.2]|S[table-format=2.2]|}
\hline
Method & {GSM8K} & {MLQA} & {Avg} \\ \hline \hline
RDS+ & 58.91 & 51.31 & 55.11 \\
LESS & 42.53 & 38.47 & 40.50 \\
\hline
\method{} (hidden states) & {\textbf{61.56}} & 51.05 & 56.31 \\
\method{} (gradients) & 52.77 & {\underline{60.19}} & {\underline{56.48}} \\
\method{} (fused) & {\underline{59.14}} & {\textbf{60.61}} & {\textbf{59.88}} \\
\hline
\end{tabular}
\end{table}

\textbf{Main results.}
The top half of Table~\ref{tab:main-results} shows the main results across the five tasks, where data are selected by Qwen3-1.7B and the performance is obtained by finetuning the same Qwen3-1.7B model.
As seen, random selection remains a strong baseline~\citep{diddee2024chasing, ivison2025large}, and is particularly competitive on OASST1, likely because the T\"ulu 2 pool is rich in instruction-tuning data.
Loss-based methods, Top-PPL and Mid-PPL, generally underperform random selection.
This is because Top-PPL favors surprising samples that often have very short answers; while this benefits the short-span F1 on DROP, it degrades the chain-of-thought reasoning required for GSM8K.
Mid-PPL is less extreme and thus maintains an overall edge over Top-PPL.
In contrast, RDS+~\citet{ivison2025large} outperforms both random and loss-based selection, whereas LESS~\citep{xia2024less} barely improves over random on average and even falls behind it on MLQA, as its max-similarity selection suffers from extreme mode seeking although Table~\ref{tab:feature-space} shows its gradient-based features are informative.
Our \method{} performs consistently well, achieving the best results on four of the five tasks and on average.
The only exception is DROP, where it ranks second to Top-PPL because \method{} tends to generate more verbose answers that are penalized by the F1 score.

\textbf{MLQA per-language results.}
\begin{table}[t]
\centering
\caption{Per-language F1 on MLQA using Qwen3-1.7B with 10k selected samples.
The average matches the MLQA column of Table~\ref{tab:main-results}.
The best and second-best results are \textbf{bold} and \underline{underlined}, respectively.}
\label{tab:mlqa-per-language}
\small
\setlength{\tabcolsep}{4pt}
\begin{tabular}{|l|S[table-format=2.2]S[table-format=2.2]S[table-format=2.2]S[table-format=2.2]S[table-format=2.2]S[table-format=2.2]S[table-format=2.2]|S[table-format=2.2]|}
\hline
Method & {Arabic} & {German} & {English} & {Spanish} & {Hindi} & {Vietnamese} & {Chinese} & {Avg} \\ \hline \hline
Random & {\underline{49.18}} & 46.61 & 59.87 & 44.13 & 35.59 & 53.37 & 49.11 & 48.27 \\
Top-PPL~\citep{ankner2024perplexed} & 46.89 & 48.76 & 63.65 & 50.73 & {\textbf{44.54}} & {\underline{54.90}} & {\underline{51.97}} & {\underline{51.63}} \\
Mid-PPL~\citep{marion2023less} & 38.98 & 14.39 & 14.60 & 16.20 & 25.76 & 19.70 & 19.47 & 21.30 \\
LESS~\citep{xia2024less} & 47.06 & 38.85 & 35.83 & 26.98 & 33.31 & 48.53 & 38.75 & 38.47 \\
RDS+~\citep{ivison2025large} & 47.56 & {\underline{51.57}} & {\underline{65.71}} & {\underline{52.23}} & {\underline{38.89}} & 52.81 & 50.39 & 51.31 \\
\hline
\method{} (ours) & {\textbf{56.82}} & {\textbf{61.58}} & {\textbf{75.85}} & {\textbf{65.57}} & 36.48 & {\textbf{65.05}} & {\textbf{62.90}} & {\textbf{60.61}} \\
\hline
\end{tabular}
\end{table}

Table~\ref{tab:mlqa-per-language} presents the breakdown results for MLQA in terms of different languages.
As seen, \method{} achieves the best F1 in six of the seven languages. The exception is Hindi, for which Top-PPL perform surprisingly well, while other methods are only slightly better or even worse than random. Among other languages, our \method{} has large margins in English, Spanish, and German, showing that our gain (reported as average in Table~\ref{tab:main-results}) is not driven by a few languages.

\textbf{Transfer to larger models.}
We further test whether the selected data transfer to larger models, where data are selected with Qwen3-1.7B and fine-tuned on Qwen3-4B and 8B (bottom half of Table~\ref{tab:main-results}). This mimics a practical application scenario, where a small model is used to select high-quality data for training large models.

Since Top-PPL and Mid-PPL underperform random in the skill-based selection setting, we drop them from now on to save compute.
As seen, \method{} outperforms both RDS+ and LESS on average and achieves the highest performance in 8 out of 10 settings. We perform a statistical test against RDS+, which is the strongest baseline. Including the 1.7B results, \method{} outperforms RDS+ under 13 of the 15 setups, which is statistically significant under a two-sided binomial test ($p \approx 0.007$). 

These results indicate that a small model with \method{} can select data that are effective for significantly larger models, suggesting that the cost of data selection can be paid once with a small model and reused for larger ones.

\begin{table}[t]
\centering
\caption{Comparison of the ICL and Train--Val proxies against test performance.
Train--Val involves fine-tuning on selected data followed by dev-set evaluation.
The final row shows the Spearman's $\rho$ between each proxy and the test score across the six methods.}
\label{tab:icl-proxy}
\small
\setlength{\tabcolsep}{5pt}
\begin{tabular}{|l|S[table-format=-1.2]S[table-format=2.2]|S[table-format=2.2]|S[table-format=-1.2]S[table-format=2.2]|S[table-format=2.2]|}
\hline
\multirow{3}{*}{Method} & \multicolumn{3}{c|}{GSM8K} & \multicolumn{3}{c|}{MLQA} \\ \cline{2-7}
 & \multicolumn{2}{c|}{Proxy} & {\multirow{2}{*}{Test}} & \multicolumn{2}{c|}{Proxy} & {\multirow{2}{*}{Test}} \\ \cline{2-3} \cline{5-6}
 & {ICL (ours)} & {Train--Val} & & {ICL (ours)} & {Train--Val} & \\ \hline \hline
Random & -3.76 & 46.51 & 40.33 & -6.93 & 46.39 & 48.27 \\
Top-PPL & -3.86 & 16.21 & 13.57 & -4.93 & 52.25 & {\underline{51.63}} \\
Mid-PPL & -3.61 & 48.66 & 42.08 & -7.31 & 21.54 & 21.30 \\
LESS~\citep{xia2024less} & -3.25 & 47.81 & 42.53 & -7.93 & 37.53 & 38.47 \\
RDS+~\citep{ivison2025large} & -2.89 & 63.08 & {\underline{58.91}} & -5.70 & 49.96 & 51.31 \\
\method{} (ours) & -2.85 & 64.50 & {\textbf{59.14}} & -3.26 & 61.11 & {\textbf{60.61}} \\
\hline
Spearman $\rho$ with test & {\textbf{1.00}} & 0.94 & {--} & 0.94 & {\textbf{1.00}} & {--} \\ \hline
\end{tabular}
\end{table}

\begin{table}[t]
\centering
\caption{Efficiency comparison.
Times are in minutes, with inference measured on 1K samples.
Speedup is the total time of training--validation divided by that of each method, computed from unrounded times.
$\rho$ denotes the Spearman correlation relative to the rankings in Table~\ref{tab:main-results}.
Experiments use an NVIDIA L40S GPU.}
\label{tab:efficiency}
\small
\setlength{\tabcolsep}{4pt}
\begin{tabular}{|l|S[table-format=2.1]S[table-format=3.1]S[table-format=3.1]r@{$\times$\hspace{8pt}}S[table-format=1.2]|S[table-format=2.1]S[table-format=3.1]S[table-format=3.1]r@{$\times$\hspace{8pt}}S[table-format=1.2]|}
\hline
\multirow{2}{*}{Method} & \multicolumn{5}{c|}{GSM8K} & \multicolumn{5}{c|}{MLQA} \\ \cline{2-11}
 & {Train} & {Infer} & {Total} & \multicolumn{1}{c}{Speedup} & {$\rho$} & {Train} & {Infer} & {Total} & \multicolumn{1}{c}{Speedup} & {$\rho$} \\ \hline \hline
ICL (ours), $M=1$ & {--} & 0.4 & 0.4 & 157 & 1.00 & {--} & 0.5 & 0.5 & 68 & 0.89 \\
ICL (ours), $M=5$ & {--} & 1.9 & 1.9 & 31 & 1.00 & {--} & 2.5 & 2.5 & 14 & 0.94 \\
ICL (ours), $M=10$ & {--} & 3.8 & 3.8 & 16 & 1.00 & {--} & 5.0 & 5.0 & 7 & 0.94 \\
\hline
Training--validation & 33.0 & 26.1 & 59.1 & 1 & 0.94 & 30.0 & 4.3 & 34.3 & 1 & 1.00 \\
\hline
\end{tabular}
\end{table}

\textbf{Analyzing the concentration parameter.}
Table~\ref{tab:kappa-sweep} compares different selection criteria and their impact on mode-seeking and mode-averaging behaviors.
As a controlled setting, all methods use the same feature space with $\beta=0.5$.
Max-similarity, as adopted by LESS, selects samples based on their highest similarity to any dev sample, leading to extreme mode seeking.
On MLQA, for instance, the max operator latches onto purely English samples, so max-similarity is even worse than random selection.
Conversely, mean-similarity selects samples based on their overall fit across the dev set; while this can be suboptimal if the average falls into a low-density region, it nevertheless performs reasonably well on MLQA because the balanced dev set keeps the average close to each language.
Round-robin, used by RDS+, mitigates extreme mode-seeking by matching each dev sample in turn, thereby promoting diversity and performing well on MLQA.
Generally, max-similarity, mean-similarity, and round-robin perform similarly on GSM8K, where the selection should more carefully balance high-quality chain-of-thought samples with diversity.

Unlike these fixed criteria, \method{} controls the balance through the concentration parameter $\kappa$, where a small $\kappa$ behaves like mean-similarity and a large $\kappa$ behaves like max-similarity. Our \method{} selects $\kappa$ by validation based on the in-context learning proxy. For the analysis, we manually set different scores for $\kappa$ and evaluate the performance.
In particular, we initialize $\kappa$ using the median heuristic~\citep{gretton2012kernel} and tune a multiplicative factor. The results are reported in Table~\ref{tab:kappa-sweep}.
In this analysis, we also omit the diversity term (setting $\alpha=0$) across all runs to isolate the effect of $\kappa$. In this way, $\kappa=0$ resembles mean-similarity selection; if $\kappa$ approaches infinity, our method becomes max-similarity selection.

Empirical results show that $\kappa$ indeed balances mode averaging and mode seeking, where small and large $\kappa$ matches the performance of mean- and max-similarity, respectively.
On GSM8K, an intermediate $\kappa$ is optimal, as a balanced $\kappa$ strikes a good balance between mode averaging and mode seeking, allowing us to pick diverse, high-quality chain-of-thought reasoning samples.
On MLQA, mode averaging is preferred, potentially because the dev set is well-balanced across languages.
By contrast, a large $\kappa$ suffers from the same mode-seeking issue as max-similarity.

Note that Table~\ref{tab:kappa-sweep} shows results of different $\kappa$ values that are set manually. However, our \method{} uses in-context learning as proxy and automatically selects the best $\kappa$ for a task, highlighting the flexibility of our approach.

\textbf{Comparing feature spaces.}
Table~\ref{tab:feature-space} compares \method{} using different feature spaces.
As seen, \method{} variants generally outperform RDS+ and LESS regardless of the features.
Within \method{} variants, hidden states and gradients provide complementary signals that suit different tasks, with hidden states performing better on GSM8K and gradients on MLQA.
By combining both, the fused space takes advantage of both worlds and achieves the best performance overall.

\textbf{Analyzing the in-context learning proxy.} 
We compare the ICL proxy with the standard training--validation approach for hyperparameter tuning. The latter is computationally inefficient as it requires full training and inference.

We use Spearman's $\rho$ to measure the ranking correlation between the dev performance (using ICL proxy or standard training--validation) and test performance for the methods in Table~\ref{tab:main-results}. As seen in Table~\ref{tab:icl-proxy}, the ICL proxy is comparable to standard training--validation in terms of method ranking: it achieves a perfect ranking on GSM8K, while training--validation does so on MLQA.
Moreover, both methods identify \method{} as the top-performing method, consistent with the test results. This shows that the ICL proxy is an effective replacement for fine-tuning and validation.

\textbf{Analyzing time efficiency.}
We further analyze the efficiency gains of the ICL proxy over training--validation in Table~\ref{tab:efficiency}, where we vary the number of in-context example draws, i.e., $M$ in Eqn.~(\ref{eqn:icl-proxy}). We see that our ICL proxy yields significant speedup of 68--157$\times$ when $M=1$, although the proxy quality is slightly lower (shown by Spearman's $\rho$ correlation). Even with $M=10$, our method remains 7--16$\times$ faster. This is understandable, as our ICL proxy eliminates the need for training, which is often the most time-consuming component of the training--validation framework. 

\textbf{Comparing kernels.}
\begin{table}[t]
\centering
\caption{Comparison of kernels within our herding framework, all using fused hidden-state and gradient features.}
\label{tab:kernel-choice}
\small
\setlength{\tabcolsep}{4pt}
\begin{tabular}{|l|S[table-format=2.2]S[table-format=2.2]|S[table-format=2.2]|}
\hline
Kernel & {GSM8K} & {MLQA} & {Avg} \\ \hline \hline
Euclidean & 47.84 & 54.97 & 51.41 \\
Cosine & 49.58 & 55.56 & 52.57 \\
RBF & {\underline{58.91}} & {\underline{60.04}} & {\underline{59.48}} \\
\hline
vMF (ours) & {\textbf{59.14}} & {\textbf{60.61}} & {\textbf{59.88}} \\
\hline
\end{tabular}
\end{table}

We compare the choice of kernel function in Table~\ref{tab:kernel-choice}.
Between Euclidean and cosine kernels, we observe that cosine similarity generally performs better. This is consistent with existing literature, which prefers using direction rather than magnitude for high-dimensional embeddings~\citep{banerjee2005clustering, reimers2019sentence, gao2021simcse}. Both RBF and vMF kernels offer an exponential decay when samples are moved apart, which enables a balance between mode averaging and mode seeking; they outperform the Euclidean and cosine kernels by a large margin. Furthermore, the angular similarity of the vMF kernel outperforms Euclidean distance, being the most effective kernel across all tested options.

\textbf{Effect of the dev set size.}
\begin{table}[t]
\centering
\caption{Effect of the dev set size on \method{}.
Full denotes the entire dev set (7,473 samples for GSM8K and 3,500 for MLQA).}
\label{tab:dev-size}
\small
\setlength{\tabcolsep}{4pt}
\begin{tabular}{|l|S[table-format=2.2]S[table-format=2.2]|}
\hline
\# Dev samples & {GSM8K} & {MLQA} \\ \hline \hline
10 & {\underline{58.98}} & 60.51 \\
100 & 58.61 & {\textbf{61.23}} \\
1,000 & 58.61 & 60.52 \\
Full & {\textbf{59.14}} & {\underline{60.61}} \\
\hline
\end{tabular}
\end{table}

We study the effect of the dev set size on our \method{}.
We progressively sample 10, 100, and 1,000 dev entries. We keep the 10K training budget and tune hyperparameters on the full dev set. 
As shown in Table~\ref{tab:dev-size}, performance is nearly flat across dev set sizes, varying by less than one point on both tasks.
This is because the dev set enters the objective only through the average kernel similarity, which a few samples already estimate well. This highlights the robustness of our approach across different dev set sizes.

\section{Conclusion}
\label{sec:conclusion}

In this work, we propose \method{}, a data selection framework for LLMs based on data herding, which balances mode averaging and mode seeking with a vMF kernel fused over hidden states and gradients.
We further propose in-context learning as an efficient proxy for tuning its hyperparameters without training.
Experiments show that \method{} outperforms existing data selection methods on average, and that our proxy is a reliable and efficient replacement for fine-tuning and validation.

\FloatBarrier

\bibliographystyle{assets/plainnat}
\bibliography{references}

\beginappendix

\begin{table}[b]
\centering
\caption{Statistics of the evaluation tasks.
Dev samples define the target skill for data selection, and test samples are used for the reported results.}
\label{tab:dataset-stats}
\small
\setlength{\tabcolsep}{4pt}
\begin{tabular}{|l|l|r|r|}
\hline
Task & Skill & {Dev} & {Test} \\ \hline \hline
GSM8K & Math reasoning & 7,473 & 1,319 \\
MLQA & Multilingual QA & 3,500 & 7,000 \\
DROP & Reading comprehension & 5,541 & 9,536 \\
XSum & Summarization & 1,000 & 1,000 \\
OASST1 & Instruction following & 500 & 500 \\
\hline
\end{tabular}
\end{table}

\section{Implementation Details}
\label{app:implementation}

We use Qwen3-1.7B~\citep{yang2025qwen3} and set the training budget to 10K, i.e., selecting 10K samples for training each task.
Following \citet{ivison2025large}, we fine-tune for 2 epochs with a learning rate of 2e-5, and following \citet{xia2024less}, we employ LoRA with rank 128. We set vMF concentration $\kappa$ by validation (Section~\ref{sec:method}).

\section{Dataset Details}
\label{app:dataset-stats}

We evaluate our method on the following five tasks:

$\bullet$ \underline{GSM8K}~\citep{cobbe2021training}: grade-school math word problems, evaluated by exact match of the final answer.

$\bullet$ \underline{MLQA}~\citep{lewis2020mlqa}: extractive question answering over passages in seven languages, evaluated by F1 averaged across languages.

$\bullet$ \underline{DROP}~\citep{dua2019drop}: reading comprehension requiring discrete reasoning (e.g., counting and arithmetic), also evaluated by F1.

$\bullet$ \underline{XSum}~\citep{narayan2018dont}: one-sentence summarization of BBC articles, evaluated by \mbox{ROUGE-L}.

$\bullet$ \underline{OASST1}~\citep{kopf2023openassistant}: open-ended instruction following, evaluated by \mbox{ROUGE-L} relative to the reference response.

Table~\ref{tab:dataset-stats}  lists the number of dev and test samples for each task.
For tasks without a dedicated dev split, we use samples from the training set as the dev set.
MLQA provides no training set, so we sample 500 dev and 1,000 test samples per language from its validation and test splits, respectively, to keep both sets balanced across languages.
For XSum, we subsample 1,000 dev and 1,000 test samples for efficiency, as summarization requires reading long contexts.
For OASST1, we split 500 dev and 500 test samples from the T\"ulu 2 pool and remove all OASST1 samples from the selection pool to prevent data leakage.

\end{document}